\documentclass[10pt,twocolumn,letterpaper]{article}

\usepackage[final]{cvpr} % [review|final|pagenumbers]{cvpr}

\usepackage{graphicx, amsmath, amssymb, caption, subcaption, multirow, overpic}
\usepackage[table]{xcolor}
\usepackage{booktabs}
\usepackage[british, english, american]{babel}
\usepackage{textpos}
\usepackage{marvosym}
\usepackage{wasysym}

\usepackage[pagebackref=false, breaklinks=true, letterpaper=true, colorlinks,
            citecolor=citecolor, linkcolor=linkcolor, bookmarks=false]{hyperref}

\definecolor{citecolor}{HTML}{0071BC}
\definecolor{linkcolor}{HTML}{ED1C24}

\def\cvprPaperID{***}
\def\confName{CVPR}
\def\confYear{2022}

\newlength\savewidth\newcommand\shline{\noalign{\global\savewidth\arrayrulewidth
  \global\arrayrulewidth 1pt}\hline\noalign{\global\arrayrulewidth\savewidth}}
\newcommand{\tablestyle}[2]{\setlength{\tabcolsep}{#1}\renewcommand{\arraystretch}{#2}\centering\footnotesize}
\renewcommand{\paragraph}[1]{\vspace{1.25mm}\noindent\textbf{#1}}

\newcolumntype{x}[1]{>{\centering\arraybackslash}p{#1pt}}
\newcolumntype{y}[1]{>{\raggedright\arraybackslash}p{#1pt}}
\newcolumntype{z}[1]{>{\raggedleft\arraybackslash}p{#1pt}}

\newcolumntype{x}[1]{>{\centering\arraybackslash}p{#1pt}}

\makeatletter\renewcommand\paragraph{\@startsection{paragraph}{4}{\z@}
	{.5em \@plus1ex \@minus.2ex}{-.5em}{\normalfont\normalsize\bfseries}}\makeatother

\newcommand{\app}{\raise.17ex\hbox{$\scriptstyle\sim$}}

\definecolor{deemph}{gray}{0.6}

\definecolor{baselinecolor}{gray}{.92}

\newcommand{\authorskip}{\hspace{2.5mm}}

\usepackage{nicefrac}       % compact symbols for 1/2, etc.
\usepackage{microtype}      % microtypography

\usepackage{multirow}
\usepackage{verbatim}
\usepackage{color}
\usepackage{float}
\usepackage{enumitem}
\usepackage{booktabs}
\usepackage{tabulary,multirow,overpic,xcolor}
\usepackage{makecell}

\usepackage{bm}
\usepackage{t1enc}
\usepackage{colortbl}
\usepackage{cite}
\usepackage{lipsum}

\begin{document}
\title{A ConvNet for the 2020s}
\author{\hspace{-2ex} Zhuang Liu$^{1,2}$\thanks{Work done during an internship at Facebook AI Research.} \authorskip Hanzi Mao$^{1}$ \authorskip Chao-Yuan Wu$^{1}$ \authorskip Christoph Feichtenhofer$^{1}$ \authorskip Trevor Darrell$^{2}$ \authorskip Saining Xie$^{1}$\thanks{Corresponding author.} \\[2mm]
\hspace{-2ex}$^1$Facebook AI Research (FAIR) \quad $^2$UC Berkeley}
 \maketitle
	\definecolor{convcolor}{HTML}{412F8A}
	\definecolor{resnetcolor}{HTML}{8DA0CB}
	\definecolor{vitcolor}{HTML}{fc8e62}

	\newcommand{\convcolor}[1]{\textcolor{convcolor}{#1}}
	\newcommand{\vitcolor}[1]{\textcolor{vitcolor}{#1}}
	\newcommand{\cnn}{ConvNeXt}

	\newcommand{\vb}{\vitcolor{$\mathbf{\circ}$\,}}
    \newcommand{\cb}{\convcolor{$\bullet$\,}}
    \newcommand{\gr}{\rowcolor[gray]{.95}}

%%%%%%%%%%%%%%%%%%%%%%%%%%%%%%%%%%%%%%%%%%%%%%%%%%%%%%%%%%%%%%%%%%%%%%%%%%%%%%%%%%%%%%%%%%%%%%%%%%%
\begin{abstract}
The ``Roaring 20s'' of visual recognition began with the introduction of Vision Transformers (ViTs), which quickly superseded ConvNets as the state-of-the-art image classification model. A vanilla ViT, on the other hand, faces difficulties when applied to general computer vision tasks such as object detection and semantic segmentation. It is the hierarchical Transformers (e.g., Swin Transformers) that reintroduced several ConvNet priors, making Transformers practically viable as a generic vision backbone and demonstrating remarkable performance on a wide variety of vision tasks. However, the effectiveness of such hybrid approaches is still largely credited to the intrinsic superiority of Transformers, rather than the inherent inductive biases of convolutions. In this work, we reexamine the design spaces and test the limits of what a pure ConvNet can achieve. We gradually ``modernize'' a standard ResNet toward the design of a vision Transformer, and discover several key components that contribute to the performance difference along the way. The outcome of this exploration is a family of pure ConvNet models dubbed ConvNeXt. Constructed entirely from standard ConvNet modules, ConvNeXts compete favorably with Transformers in terms of accuracy and scalability, achieving 87.8\% ImageNet top-1 accuracy and outperforming Swin Transformers on COCO detection and ADE20K segmentation, while maintaining the simplicity and efficiency of standard ConvNets. 
\end{abstract}
%##################################################################################################

\begin{textblock*}{.8\textwidth}[.5,0](0.53\textwidth, -.723\textwidth)
\centering
{\hspace{-6ex} \small Code: \url{https://github.com/facebookresearch/ConvNeXt}}
\end{textblock*}

%%%%%%%%%%%%%%%%%%%%%%%%%%%%%%%%%%%%%%%%%%%%%%%%%%%%%%%%%%%%%%%%%%%%%%%%%%%%%%%%%%%%%%%%%%%%%%%%%%%
\section{Introduction}
\label{sec:intro}

Looking back at the 2010s, the decade was marked by the monumental progress and impact of deep learning. The primary driver was the renaissance of neural networks, particularly convolutional neural networks (ConvNets). Through the decade, the field of visual recognition successfully shifted from engineering features to designing (ConvNet) architectures. Although the invention of back-propagation-trained ConvNets dates all the way back to the 1980s~\cite{LeCun1989}, it was not until late 2012 that we saw its true potential for visual feature learning. The introduction of AlexNet~\cite{Krizhevsky2012} precipitated the ``ImageNet moment''~\cite{Russakovsky2015}, ushering in a new era of computer vision. The field has since evolved at a rapid speed. Representative ConvNets like VGGNet~\cite{Simonyan2014}, Inceptions~\cite{Szegedy2015}, ResNe(X)t~\cite{He2016,Xie2017}, DenseNet~\cite{Huang2017}, MobileNet~\cite{Howard2017}, EfficientNet~\cite{Tan2019efficientnet} and RegNet~\cite{Radosavovic2020designing} focused on different aspects of accuracy, efficiency and scalability, and popularized many useful design principles. 

\begin{figure}[t]
\centering
\vspace{-0em}
\includegraphics[width=0.48\textwidth]{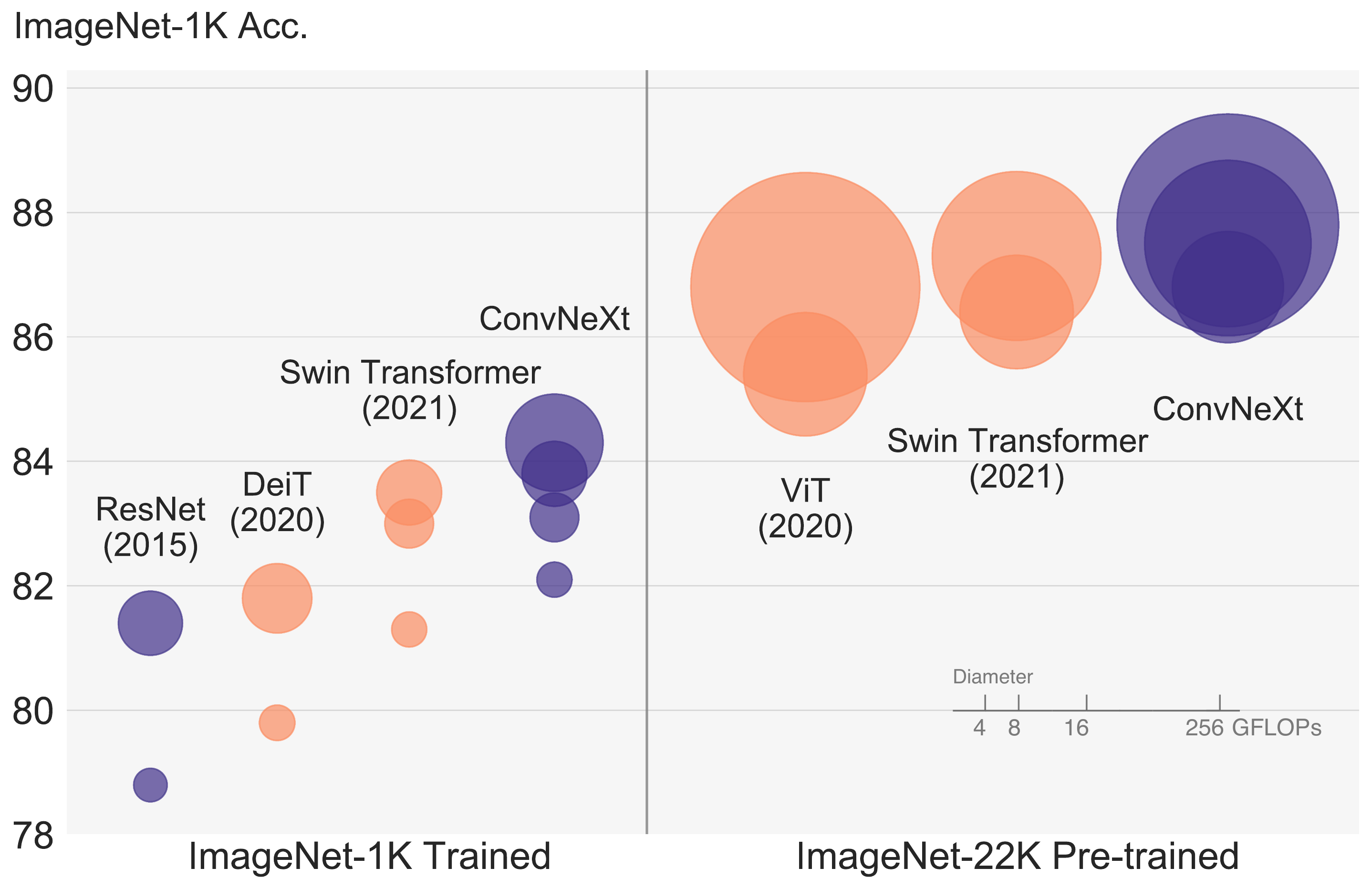}
\vspace{-1.3em}
\caption{\textbf{ImageNet-1K classification} results for \cb ConvNets and \vb vision Transformers. Each bubble's area is proportional to FLOPs of a variant in a model family. ImageNet-1K/22K models here take 224$^2$/384$^2$ images respectively. ResNet and ViT results were obtained with improved training procedures over the original papers. We demonstrate that a standard ConvNet model can achieve the same level of scalability as hierarchical vision Transformers while being much simpler in design.}
\label{fig:teaser}\vspace{-1.7em}
\end{figure}

The full dominance of ConvNets in computer vision was not a coincidence: in many application scenarios, a ``sliding window'' strategy is intrinsic to visual processing, particularly when working with high-resolution images. ConvNets have several built-in inductive biases that make them well-suited to a wide variety of computer vision applications. The most important one is translation equivariance, which is a desirable property for tasks like objection detection. ConvNets are also inherently efficient due to the fact that when used in a sliding-window manner, the computations are shared~\cite{Sermanet2014}. For many decades, this has been the default use of ConvNets, generally on limited object categories such as digits \cite{lecun1998gradient}, faces \cite{vaillant1994original,rowley1998neural} and pedestrians \cite{sermanet2013pedestrian, dollar2010fastest}. Entering the 2010s, the region-based detectors \cite{Girshick2014, Girshick2015, Ren2015, He2017} further elevated ConvNets to the position of being the fundamental building block in a visual recognition system. 

Around the same time, the odyssey of neural network design for natural language processing (NLP) took a very different path, as the Transformers replaced recurrent neural networks to become the dominant backbone architecture. Despite the disparity in the task of interest between language and vision domains, the two streams surprisingly converged in the year 2020, as the introduction of Vision Transformers (ViT) completely altered the landscape of network architecture design. Except for the initial ``patchify'' layer, which splits an image into a sequence of patches, ViT introduces no image-specific inductive bias and makes minimal changes to the original NLP Transformers. One primary focus of ViT is on the scaling behavior: with the help of larger model and dataset sizes, Transformers can outperform standard ResNets by a significant margin. Those results on image classification tasks are inspiring, but computer vision is not limited to image classification. As discussed previously, solutions to numerous computer vision tasks in the past decade depended significantly on a sliding-window, fully-convolutional paradigm. Without the ConvNet inductive biases, a vanilla ViT model faces many challenges in being adopted as a generic vision backbone. The biggest challenge is ViT's global attention design, which has a quadratic complexity with respect to the input size. This might be acceptable for ImageNet classification, but quickly becomes intractable with higher-resolution inputs. 

Hierarchical Transformers employ a hybrid approach to bridge this gap. For example, the ``sliding window'' strategy (\eg attention within local windows) was reintroduced to Transformers, allowing them to behave more similarly to ConvNets. Swin Transformer~\cite{Liu2021swin} is a milestone work in this direction, demonstrating for the first time that Transformers can be adopted as a generic vision backbone and achieve state-of-the-art performance across a range of computer vision tasks beyond image classification. Swin Transformer's success and rapid adoption also revealed one thing: the essence of convolution is not becoming irrelevant; rather, it remains much desired and has never faded.

Under this perspective, many of the advancements of Transformers for computer vision have been aimed at bringing back convolutions. These attempts, however, come at a cost: a naive implementation of sliding window self-attention can be expensive~\cite{ramachandran2019stand}; with advanced approaches such as cyclic shifting~\cite{Liu2021swin}, the speed can be optimized but the system becomes more sophisticated in design. 
On the other hand, it is almost ironic that a ConvNet already satisfies many of those desired properties, albeit in a straightforward, no-frills way. The only reason ConvNets appear to be losing steam is that (hierarchical) Transformers surpass them in many vision tasks, and the performance difference is usually attributed to the superior scaling behavior of Transformers, with multi-head self-attention being the key component. 

Unlike ConvNets, which have progressively improved over the last decade, the adoption of Vision Transformers was a step change. In recent literature, system-level comparisons (\eg a Swin Transformer \vs a ResNet) are usually adopted when comparing the two. ConvNets and hierarchical vision Transformers become different and similar at the same time: they are both equipped with similar inductive biases, but differ significantly in the training procedure and macro/micro-level architecture design. 
In this work, 
we investigate the architectural distinctions between ConvNets and Transformers and try to identify the confounding variables when comparing the network performance.
Our research is intended to bridge the gap between the pre-ViT and post-ViT eras for ConvNets, as well as to test the limits of what a pure ConvNet can achieve. 

To do this, we start with a standard ResNet (\eg ResNet-50) trained with an improved procedure. We gradually ``modernize'' the architecture to the construction of a hierarchical vision Transformer (\eg Swin-T). Our exploration is directed by a key question: \emph{How do design decisions in Transformers impact ConvNets' performance?} We discover several key components that contribute to the performance difference along the way. As a result, we propose a family of \textit{pure ConvNets} dubbed \cnn{}. 
We evaluate \cnn{s} on a variety of vision tasks such as ImageNet classification~\cite{Deng2009}, object detection/segmentation on COCO\cite{Lin2014}, and semantic segmentation on ADE20K~\cite{Zhou2019}. Surprisingly, \cnn{s}, constructed entirely from standard ConvNet modules, compete favorably with Transformers in terms of accuracy, scalability and robustness across all major benchmarks. 
ConvNeXt maintains the efficiency of standard ConvNets, and the fully-convolutional nature for both training and testing makes it extremely simple to implement.

We hope the new observations and discussions can challenge some common beliefs and encourage people to rethink the importance of convolutions in computer vision.

%##################################################################################################
\begin{figure}[t] 
% \vspace{-0.5em}
\centering
\includegraphics[width=0.46\textwidth]{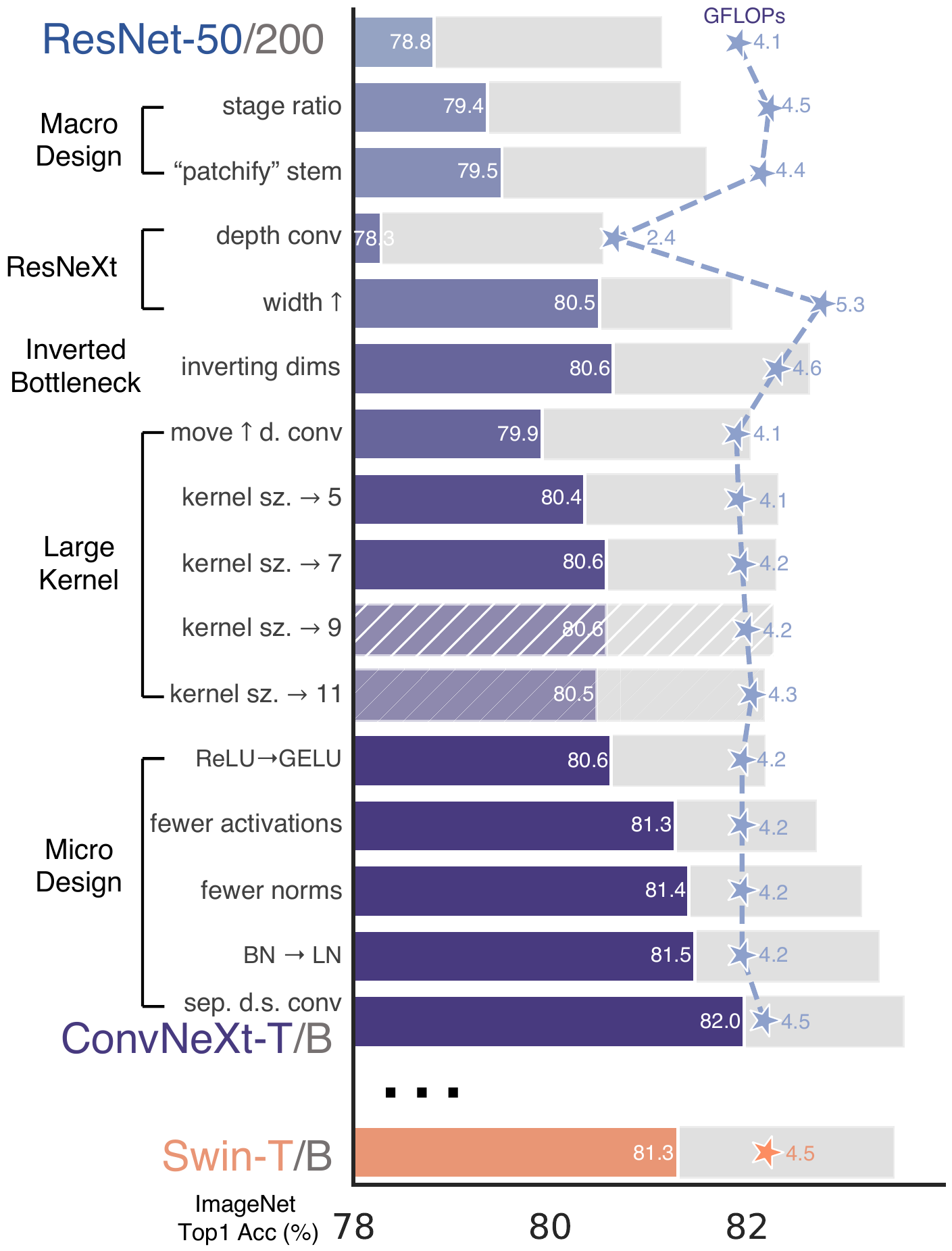}
\vspace{-.8em}
\caption{We modernize a standard ConvNet (ResNet) towards the design of a hierarchical vision Transformer (Swin), without introducing any attention-based modules. The foreground bars are model accuracies in the ResNet-50/Swin-T FLOP regime; results for the ResNet-200/Swin-B regime are shown with the gray bars. A hatched bar means the modification is not adopted. Detailed results for both regimes are in the appendix. Many Transformer architectural choices can be incorporated in a ConvNet, and they lead to increasingly better performance. In the end, our pure ConvNet model, named ConvNeXt, can outperform the Swin Transformer.}
\label{fig:morph_main}\vspace{-1.5em}
\end{figure}
%##################################################################################################

\section{Modernizing a ConvNet: a Roadmap}
\label{sec:modernizing}
In this section, we provide a trajectory going from a ResNet to a ConvNet that bears a resemblance to Transformers. We consider two model sizes in terms of FLOPs, one is the ResNet-50 / Swin-T regime with FLOPs around $4.5\times 10^9$ and the other being ResNet-200 / Swin-B regime which has FLOPs around $15.0\times 10^9$. For simplicity, we will present the results with the ResNet-50 / Swin-T complexity models. The conclusions for higher capacity models are consistent and results can be found in Appendix~\ref{sec:modernizing_result}.

At a high level, our explorations are directed to investigate and follow different levels of designs from a Swin Transformer while maintaining the network's simplicity as a standard ConvNet. The roadmap of our exploration is as follows. Our starting point is a ResNet-50 model. We first train it with similar training techniques used to train vision Transformers and obtain much improved results compared to the original ResNet-50. This will be our baseline. We then study a series of design decisions which we summarized as 1) macro design, 2) ResNeXt, 3) inverted bottleneck, 4) large kernel size, and 5) various layer-wise micro designs. In Figure~\ref{fig:morph_main}, we show the procedure and the results we are able to achieve with each step of the ``network modernization''. Since network complexity is closely correlated with the final performance, the FLOPs are roughly controlled over the course of the exploration, though at intermediate steps the FLOPs might be higher or lower than the reference models. All models are trained and evaluated on ImageNet-1K.

\subsection{Training Techniques}
Apart from the design of the network architecture, the training procedure also affects the ultimate performance. 
Not only did vision Transformers bring a new set of modules and architectural design decisions, but they also introduced different training techniques (\eg AdamW optimizer) to vision.
This pertains mostly to the optimization strategy and associated hyper-parameter settings.
Thus, the first step of our exploration is to train a baseline model with the vision Transformer training procedure, in this case, ResNet-50/200. 
Recent studies~\cite{bello2021revisiting, Wightman2021resnet} demonstrate that a set of modern training techniques can significantly enhance the performance of a simple ResNet-50 model. In our study, we use a training recipe that is close to DeiT's~\cite{Touvron2020} and Swin Transformer's~\cite{Liu2021swin}. The training is extended to 300 epochs from the original 90 epochs for ResNets. We use the AdamW optimizer~\cite{Loshchilov2019}, data augmentation techniques such as Mixup~\cite{Zhang2018a}, Cutmix~\cite{Yun2019}, RandAugment~\cite{Cubuk2020}, Random Erasing~\cite{Zhong2020}, and regularization schemes including Stochastic Depth~\cite{Huang2017} and Label Smoothing~\cite{Szegedy2016a}. The complete set of hyper-parameters we use can be found in Appendix~\ref{subsec:setting}. By itself, this enhanced training recipe increased the performance of the ResNet-50 model from 76.1\% \cite{torchvision} to 78.8\% (+2.7\%), implying that a significant portion of the performance difference between traditional ConvNets and vision Transformers may be due to the training techniques. We will use this fixed training recipe with the same hyperparameters throughout the ``modernization'' process. Each reported accuracy on the ResNet-50 regime is an average obtained from training with three different random seeds.

\subsection{Macro Design}
We now analyze Swin Transformers' macro network design. Swin Transformers follow ConvNets~\cite{Simonyan2015, He2016} to use a multi-stage design, where each stage has a different feature map resolution. There are two interesting design considerations: the stage compute ratio, and the ``stem cell'' structure.
\paragraph{Changing stage compute ratio.} 
The original design of the computation distribution across stages in ResNet was largely empirical. The heavy ``res4'' stage was meant to be compatible with downstream tasks like object detection, where a detector head operates on the 14$\times$14 feature plane. 
Swin-T, on the other hand, followed the same principle but with a slightly different stage compute ratio of 1:1:3:1. For larger Swin Transformers, the ratio is 1:1:9:1. Following the design, we adjust the number of blocks in each stage from (3, 4, 6, 3) in ResNet-50 to (3, 3, 9, 3), which also aligns the FLOPs with Swin-T. This improves the model accuracy from 78.8\% to 79.4\%. Notably, researchers have thoroughly investigated the distribution of computation~\cite{Radosavovic2019network, Radosavovic2020designing}, and a more optimal design is likely to exist. 

\textit{From now on, we will use this stage compute ratio.} 

\paragraph{Changing stem to ``Patchify''.}
Typically, the stem cell design is concerned with how the input images will be processed at the network's beginning. Due to the redundancy inherent in natural images, a common stem cell will aggressively downsample the input images to an appropriate feature map size in both standard ConvNets and vision Transformers. The stem cell in standard ResNet contains a 7$\times$7 convolution layer with stride 2, followed by a max pool, which results in a 4$\times$ downsampling of the input images. In vision Transformers, a more aggressive ``patchify'' strategy is used as the stem cell, which corresponds to a large kernel size (e.g. kernel size = 14 or 16) and non-overlapping convolution. Swin Transformer uses a similar ``patchify'' layer, but with a smaller patch size of 4 to accommodate the architecture's multi-stage design. 
We replace the ResNet-style stem cell with a patchify layer implemented using a 4$\times$4, stride 4 convolutional layer. The accuracy has changed from 79.4\% to 79.5\%. This suggests that the stem cell in a ResNet may be substituted with a simpler ``patchify'' layer à la ViT which will result in similar performance.

\textit{We will use the ``patchify stem'' (4$\times$4 non-overlapping convolution) in the network.} 

\subsection{ResNeXt-ify}
\vspace{-0.2em} 
In this part, we attempt to adopt the idea of ResNeXt~\cite{Xie2017}, which has a better FLOPs/accuracy trade-off than a vanilla ResNet. The core component is grouped convolution, where the convolutional filters are separated into different groups. At a high level, ResNeXt's guiding principle is to ``use more groups, expand width''. More precisely, ResNeXt employs grouped convolution for the 3$\times$3 conv layer in a bottleneck block. As this significantly reduces the FLOPs, the network width is expanded to compensate for the capacity loss. 

In our case we use depthwise convolution, a special case of grouped convolution where the number of groups equals the number of channels. Depthwise conv has been popularized by MobileNet~\cite{Howard2017} and Xception~\cite{Chollet2017}. We note that depthwise convolution is similar to the weighted sum operation in self-attention, which operates on a per-channel basis, \ie, only mixing information in the spatial dimension. The combination of depthwise conv and $1\times1$ convs leads to a separation of spatial and channel mixing, a property shared by vision Transformers, where each operation either mixes information across spatial or channel dimension, but not both. The use of depthwise convolution effectively reduces the network FLOPs and, as expected, the accuracy. Following the strategy proposed in ResNeXt, we increase the network width to the same number of channels as Swin-T's (from 64 to 96). This brings the network performance to 80.5\% with increased FLOPs (5.3G).
  
\textit{We will now employ the ResNeXt design.}

\subsection{Inverted Bottleneck}
\vspace{-0.2em}
One important design in every Transformer block is that it creates an inverted bottleneck, \ie, the hidden dimension of the MLP block is four times wider than the input dimension (see Figure~\ref{fig:block}). Interestingly, this Transformer design is connected to the inverted bottleneck design with an expansion ratio of 4 used in ConvNets. The idea was popularized by MobileNetV2~\cite{Sandler2018}, and has subsequently gained traction in several advanced ConvNet architectures~\cite{Tan2019efficientnet,tan2019mnasnet}.

Here we explore the inverted bottleneck design. Figure~\ref{fig:inverted} (a) to (b) illustrate the configurations. Despite the increased FLOPs for the depthwise convolution layer, this change reduces the whole network FLOPs to 4.6G, due to the significant FLOPs reduction in the downsampling residual blocks' shortcut 1$\times$1 conv layer.
Interestingly, this results in slightly improved performance (80.5\% to 80.6\%). In the ResNet-200 / Swin-B regime, this step brings even more gain (81.9\% to 82.6\%) also with reduced FLOPs.

\textit{We will now use inverted bottlenecks.}

\subsection{Large Kernel Sizes}
\vspace{-0.2em}

In this part of the exploration, we focus on the behavior of large convolutional kernels. One of the most distinguishing aspects of vision Transformers is their non-local self-attention, which enables each layer to have a global receptive field. While large kernel sizes have been used in the past with ConvNets~\cite{Krizhevsky2012,Szegedy2015}, the gold standard (popularized by VGGNet~\cite{Simonyan2015}) is to stack small kernel-sized (3$\times$3) conv layers, which have efficient hardware implementations on modern GPUs~\cite{Lavin2016FastAF}. Although Swin Transformers reintroduced the local window to the self-attention block, the window size is at least 7$\times$7, significantly larger than the ResNe(X)t kernel size of 3$\times$3. Here we revisit the use of large kernel-sized convolutions for ConvNets. 

\paragraph{Moving up depthwise conv layer.}

\begin{figure}
\centering
\includegraphics[width=0.48\textwidth]{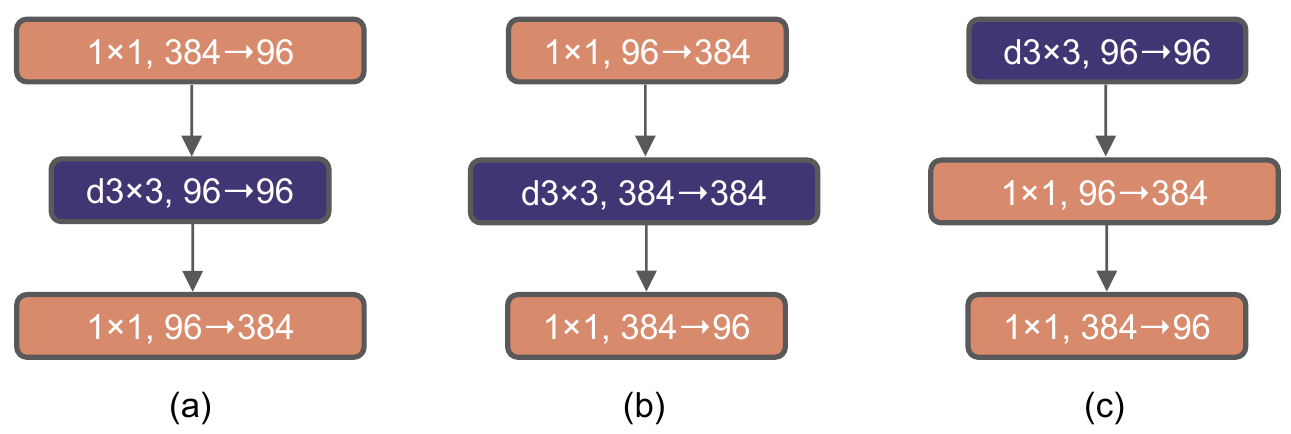}
\vspace{-1.5em}
\caption{\textbf{Block modifications and resulted specifications.} \textbf{(a)} is a ResNeXt block; in \textbf{(b)} we create an inverted bottleneck block and in \textbf{(c)} the position of the spatial depthwise conv layer is moved up.}
\label{fig:inverted}\vspace{-1em}
\end{figure}

 To explore large kernels, one prerequisite is to move up the position of the depthwise conv layer (Figure \ref{fig:inverted} (b) to (c)). That is a design decision also evident in Transformers: the MSA block is placed prior to the MLP layers. As we have an inverted bottleneck block, this is a natural design choice --- the complex/inefficient modules (MSA, large-kernel conv) will have fewer channels, while the efficient, dense 1$\times$1 layers will do the heavy lifting. This intermediate step reduces the FLOPs to 4.1G, resulting in a temporary performance degradation to 79.9\%.

\paragraph{Increasing the kernel size.}
With all of these preparations, the benefit of adopting larger kernel-sized convolutions is significant. We experimented with several kernel sizes, including 3, 5, 7, 9, and 11. The network's performance increases from 79.9\% (3$\times$3) to 80.6\% (7$\times$7), while the network's FLOPs stay roughly the same. Additionally, we observe that the benefit of larger kernel sizes reaches a saturation point at 7$\times$7. We verified this behavior in the large capacity model too: a ResNet-200 regime model does not exhibit further gain when we increase the kernel size beyond 7$\times$7. 

\textit{We will use 7$\times$7 depthwise conv in each block.} 

At this point, we have concluded our examination of network architectures on a macro scale. Intriguingly, a significant portion of the design choices taken in a vision Transformer may be mapped to ConvNet instantiations. 

\subsection{Micro Design}
In this section, we investigate several other architectural differences at a micro scale --- most of the explorations here are done at the layer level, focusing on specific choices of activation functions and normalization layers.  
\paragraph{Replacing ReLU with GELU}
One discrepancy between NLP and vision architectures is the specifics of which activation functions to use. Numerous activation functions have been developed over time, but the Rectified Linear Unit (ReLU)~\cite{Nair2010} is still extensively used in ConvNets due to its simplicity and efficiency. ReLU is also used as an activation function in the original Transformer paper~\cite{Vaswani2017}. The Gaussian Error Linear Unit, or GELU~\cite{Hendrycks2016}, which can be thought of as a smoother variant of ReLU, is utilized in the most advanced Transformers, including Google's BERT~\cite{Devlin2019} and OpenAI's GPT-2~\cite{Radford2019}, and, most recently, ViTs. We find that ReLU can be substituted with GELU in our ConvNet too, although the accuracy stays unchanged (80.6\%).

\begin{figure}[t]
\centering
\includegraphics[width=0.485\textwidth]{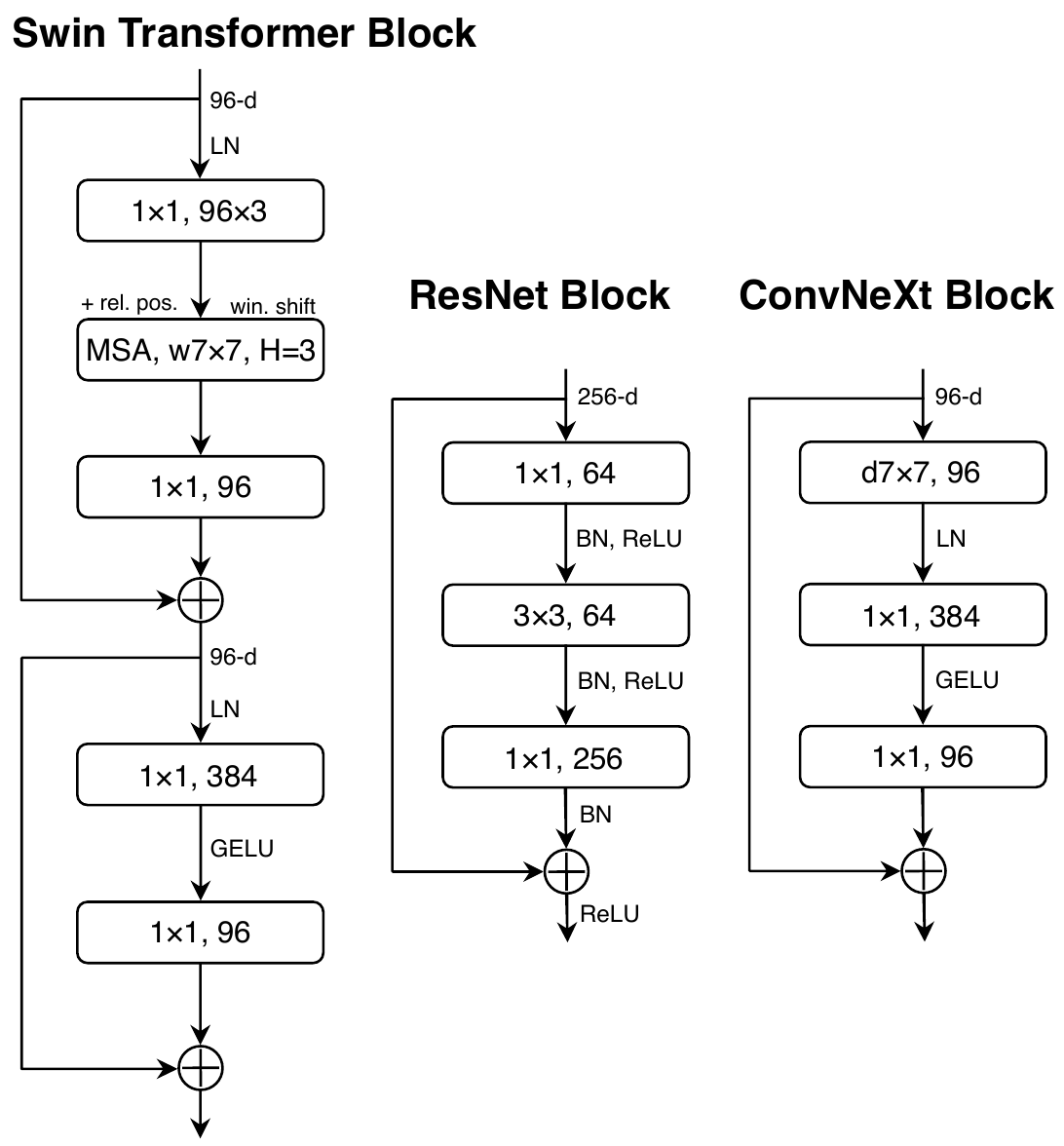}
\vspace{-2em}
\caption{\textbf{Block designs} for a ResNet, a Swin Transformer, and a \cnn. Swin Transformer's block is more sophisticated due to the presence of multiple specialized modules and two residual connections. For simplicity, we note the linear layers in Transformer MLP blocks also as ``1$\times$1 convs'' since they are equivalent.}
\label{fig:block}
\vspace{-1em}
\end{figure}

\paragraph{Fewer activation functions.}
One minor distinction between a Transformer and a ResNet block is that Transformers have fewer activation functions. Consider a Transformer block with key/query/value linear embedding layers, the projection layer, and two linear layers in an MLP block. There is only one activation function present in the MLP block. In comparison, it is common practice to append an activation function to each convolutional layer, including the $1\times1$ convs. Here we examine how performance changes when we stick to the same strategy. As depicted in Figure~\ref{fig:block}, we eliminate all GELU layers from the residual block except for one between two $1\times1$ layers, replicating the style of a Transformer block. This process improves the result by 0.7\% to 81.3\%, practically matching the performance of Swin-T.

\textit{We will now use a single GELU activation in each block.}

\paragraph{Fewer normalization layers.}
Transformer blocks usually have fewer normalization layers as well. Here we remove two BatchNorm (BN) layers, leaving only one BN layer before the conv $1\times1$ layers. This further \textit{boosts} the performance to 81.4\%, already surpassing Swin-T's result. Note that we have even fewer normalization layers per block than Transformers, as empirically we find that adding one additional BN layer at the beginning of the block does not improve the performance. 

\paragraph{Substituting BN with LN.}
BatchNorm~\cite{Ioffe2017} is an essential component in ConvNets as it improves the convergence and reduces overfitting. However, BN also has many intricacies that can have a detrimental effect on the model's performance~\cite{wu2021rethinking}. There have been numerous attempts at developing alternative normalization~\cite{Salimans2016,Ulyanov2016,Wu2018} techniques, but BN has remained the preferred option in most vision tasks.
On the other hand, the simpler Layer Normalization~\cite{Ba2016} (LN) has been used in Transformers, resulting in good performance across different application scenarios.

Directly substituting LN for BN in the original ResNet will result in suboptimal performance~\cite{Wu2018}. With all the modifications in network architecture and training techniques, here we revisit the impact of using LN in place of BN. We observe that our ConvNet model does not have any difficulties training with LN; in fact, the performance is slightly better, obtaining an accuracy of 81.5\%.

\textit{From now on, we will use one LayerNorm as our choice of normalization in each residual block.}
 
\paragraph{Separate downsampling layers.}

In ResNet, the spatial downsampling is achieved by the residual block at the start of each stage, using 3$\times$3 conv with stride 2 (and 1$\times$1 conv with stride 2 at the shortcut connection). In Swin Transformers, a separate downsampling layer is added between stages. We explore a similar strategy in which we use 2$\times$2 conv layers with stride 2 for spatial downsampling. This modification surprisingly leads to diverged training. Further investigation shows that, adding normalization layers wherever spatial resolution is changed can help stablize training. These include several LN layers also used in Swin Transformers: one before each downsampling layer, one after the stem, and one after the final global average pooling. 
We can improve the accuracy to 82.0\%, significantly exceeding Swin-T's 81.3\%.

\textit{
We will use separate downsampling layers. This brings us to our final model, which we have dubbed \cnn{}.}

\textit{A comparison of ResNet, Swin, and \cnn{} block structures can be found in Figure \ref{fig:block}. A comparison of ResNet-50, Swin-T and \cnn{}-T's detailed architecture specifications can be found in Table~\ref{table:arch-spec}.}

\paragraph{Closing remarks.}
We have finished our first ``playthrough'' and discovered \cnn{}, a pure ConvNet, that can outperform the Swin Transformer for ImageNet-1K classification in this compute regime. 
It is worth noting that \emph{all design choices discussed so far are adapted from vision Transformers. \emph{In addition,} these designs are not novel even in the ConvNet literature --- they have all been researched separately, but not collectively, over the last decade.}
Our \cnn{} model has approximately the same FLOPs, \#params., throughput, and memory use as the Swin Transformer, but does not require specialized modules such as shifted window attention or relative position biases. 

These findings are encouraging but not yet completely convincing --- our exploration thus far has been limited to a small scale, but vision Transformers' scaling behavior is what truly distinguishes them. Additionally, the question of whether a ConvNet can compete with Swin Transformers on downstream tasks such as object detection and semantic segmentation is a central concern for computer vision practitioners.
In the next section, we will scale up our \cnn{} models both in terms of data and model size, and evaluate them on a diverse set of visual recognition tasks.

\section{Empirical Evaluations on ImageNet}
\label{sec:convnext_config}
 We construct different \cnn{} variants, \cnn{}-T/S/B/L, to be of similar complexities to Swin-T/S/B/L \cite{Liu2021swin}. \cnn{}-T/B is the end product of the ``modernizing'' procedure on ResNet-50/200 regime, respectively. In addition, we build a larger \cnn{}-XL to further test the scalability of \cnn{}. The variants only differ in the number of channels $C$, and the number of blocks $B$ in each stage. Following both ResNets and Swin Transformers, the number of channels doubles at each new stage. We summarize the configurations below:
\begin{itemize}[leftmargin=-.1ex]
\setlength\itemsep{-.2em}
\small{
\item[]\cb\cnn{-T}: $C=(96,192,384,768)$, $B=(3,3,9,3)$
\item[]\cb\cnn{-S}: $C=(96,192,384,768)$, $B=(3,3,27,3)$
\item[]\cb\cnn{-B}: $C=(128,256,512,1024)$, $B=(3,3,27,3)$
\item[]\cb\cnn{-L}: $C=(192,384,768,1536)$, $B=(3,3,27,3)$
\item[]\cb\cnn{-XL}: $C=(256,512,1024,2048)$, $B=(3,3,27,3)$
}
\end{itemize}

\subsection{Settings}
The ImageNet-1K dataset consists of 1000 object classes with 1.2M training images. We report ImageNet-1K top-1 accuracy on the validation set. We also conduct pre-training on ImageNet-22K, a larger dataset of 21841 classes (a superset of the 1000 ImageNet-1K classes) with $\sim$14M images for pre-training, and then fine-tune the pre-trained model on ImageNet-1K for evaluation.  We summarize our training setups below. More details can be found in Appendix~\ref{sec:setting}. 

\paragraph{Training on ImageNet-1K.} We train \cnn{}s for 300 epochs using AdamW~\cite{Loshchilov2019} with a learning rate of 4e-3. There is a 20-epoch linear warmup and a cosine decaying schedule afterward. We use a batch size of 4096 and a weight decay of 0.05. For data augmentations, we adopt common schemes including Mixup~\cite{Zhang2018a}, Cutmix~\cite{Yun2019}, RandAugment~\cite{Cubuk2020}, and Random Erasing~\cite{Zhong2020}. We regularize the networks with Stochastic Depth~\cite{Huang2016deep} and Label Smoothing~\cite{Szegedy2016a}. Layer Scale~\cite{Touvron2021GoingDW} of initial value 1e-6 is applied. We use Exponential Moving Average (EMA)~\cite{Polyak1992} as we find it alleviates larger models' overfitting.

\paragraph{Pre-training on ImageNet-22K.} We pre-train \cnn{}s on ImageNet-22K for 90 epochs with a warmup of 5 epochs. We do not use EMA. Other settings follow ImageNet-1K.

\paragraph{Fine-tuning on ImageNet-1K.}
We fine-tune ImageNet-22K pre-trained models on ImageNet-1K for 30 epochs. We use AdamW, a learning rate of 5e-5, cosine learning rate schedule, layer-wise learning rate decay~\cite{Clark2020,Bao2021}, no warmup, a batch size of 512, and weight decay of 1e-8. 
The default pre-training, fine-tuning, and testing resolution is 224$^2$. Additionally, we fine-tune at a larger resolution of 384$^2$, for both ImageNet-22K and ImageNet-1K pre-trained models. 

Compared with ViTs/Swin Transformers, \cnn{}s are simpler to fine-tune at different resolutions, as the network is fully-convolutional and there is no need to adjust the input patch size or interpolate absolute/relative position biases.

%##################################################################################################
\begin{table}[t]
\centering
\small
\addtolength{\tabcolsep}{-5.pt}
\begin{tabular}{lccccc}
   model & \begin{tabular}[c]{@{}c@{}}image \\ size\end{tabular} & \#param. & FLOPs & \begin{tabular}[c]{@{}c@{}}throughput\\ (image / s)\end{tabular} & \begin{tabular}[c]{@{}c@{}} IN-1K \\ top-1 acc.\end{tabular} \\
\Xhline{1.0pt}
\multicolumn{6}{c}{\scriptsize{ImageNet-1K trained models}} \\
\cb RegNetY-16G~\cite{Radosavovic2020designing} & 224$^2$ & 84M & 16.0G & 334.7 & 82.9 \\
\cb EffNet-B7~\cite{Tan2019efficientnet} & 600$^2$ & 66M & 37.0G & 55.1 & 84.3 \\
\cb EffNetV2-L~\cite{tan2021efficientnetv2} & 480$^2$ & 120M & 53.0G & 83.7 & 85.7 \\

\hline

\vb DeiT-S~\cite{Touvron2020} & 224$^2$ & 22M & 4.6G & 978.5 & 79.8 \\ 
\vb DeiT-B~\cite{Touvron2020} & 224$^2$ & 87M & 17.6G & 302.1 & 81.8 \\
\hline
\vb Swin-T   & 224$^2$ & 28M & 4.5G & 757.9 & 81.3 \\
\gr
\cb \cnn{}-T & 224$^2$ & 29M & 4.5G & 774.7 & \textbf{82.1} \\ 

\vb Swin-S   & 224$^2$ & 50M & 8.7G & 436.7 & 83.0 \\
\gr
\cb \cnn{}-S & 224$^2$ & 50M & 8.7G & 447.1 & \textbf{83.1} \\ 

\vb Swin-B   & 224$^2$ & 88M & 15.4G& 286.6& 83.5 \\
\gr
\cb \cnn{}-B & 224$^2$ & 89M & 15.4G& 292.1& \textbf{83.8} \\ 

\vb Swin-B   & 384$^2$ & 88M & 47.1G& 85.1 & 84.5 \\
\gr
\cb \cnn{}-B & 384$^2$ & 89M & 45.0G& 95.7 & \textbf{85.1} \\

\gr
\cb \cnn{}-L & 224$^2$ & 198M & 34.4G & 146.8 & \textbf{84.3} \\
\gr
\cb \cnn{}-L & 384$^2$ & 198M & 101.0G& 50.4 & \textbf{85.5} \\
\hline

\multicolumn{6}{c}{\scriptsize{ImageNet-22K pre-trained models}}  \\  
\cb R-101x3~\cite{Kolesnikov2020} & 384$^2$ & 388M & 204.6G & - & 84.4 \\
\cb R-152x4~\cite{Kolesnikov2020} & 480$^2$ & 937M & 840.5G & - & 85.4 \\
\cb EffNetV2-L~\cite{tan2021efficientnetv2} & 480$^2$ & 120M & 53.0G & 83.7 & 86.8 \\
\cb EffNetV2-XL~\cite{tan2021efficientnetv2} & 480$^2$ & 208M & 94.0G & 56.5 & 87.3 \\

\hline
\vb ViT-B/16 (\Telefon)~\cite{steiner2021train} & 384$^2$ & 87M & 55.5G & 93.1 & 85.4 \\ 
\vb ViT-L/16 (\Telefon)~\cite{steiner2021train} & 384$^2$ & 305M & 191.1G & 28.5 & 86.8 \\

\hline

\gr
\cb \cnn{}-T & 224$^2$ & 29M  & 4.5G & 774.7 & \textbf{82.9} \\
\gr
\cb \cnn{}-T & 384$^2$ & 29M  & 13.1G & 282.8 & \textbf{84.1} \\
\gr
\cb \cnn{}-S & 224$^2$ & 50M  & 8.7G & 447.1 & \textbf{84.6} \\
\gr
\cb \cnn{}-S & 384$^2$ & 50M  & 25.5G & 163.5 & \textbf{85.8} \\

\vb Swin-B   & 224$^2$ & 88M  & 15.4G & 286.6 & 85.2          \\
\gr
\cb \cnn{}-B & 224$^2$ & 89M  & 15.4G & 292.1 & \textbf{85.8} \\

\vb Swin-B   & 384$^2$ & 88M  & 47.0G & 85.1  & 86.4          \\
\gr
\cb \cnn{}-B & 384$^2$ & 89M  & 45.1G & 95.7  & \textbf{86.8} \\

\vb Swin-L   & 224$^2$ & 197M & 34.5G & 145.0 &  86.3         \\
\gr
\cb \cnn{}-L & 224$^2$ & 198M & 34.4G & 146.8 & \textbf{86.6}\\

\vb Swin-L   & 384$^2$ & 197M & 103.9G& 46.0  & 87.3          \\
\gr
\cb \cnn{}-L & 384$^2$ & 198M & 101.0G& 50.4  & \textbf{87.5}  \\
\gr
\cb \cnn{}-XL & 224$^2$ & 350M & 60.9G & 89.3  & \textbf{87.0}  \\
\gr
\cb \cnn{}-XL & 384$^2$ & 350M & 179.0G& 30.2  & \textbf{87.8}  \\

\hline
\end{tabular}
\normalsize
\caption{\textbf{Classification accuracy on ImageNet-1K.} Similar to Transformers, \cnn{} also shows promising scaling behavior with higher-capacity models and a larger (pre-training) dataset. Inference throughput is measured on a V100 GPU, following~\cite{Liu2021swin}. On an A100 GPU, \cnn{} can have a much higher throughput than Swin Transformer. See Appendix~\ref{sec:a100}. (\Telefon)ViT results with 90-epoch AugReg~\cite{steiner2021train} training, provided through personal communication with the authors.}
\label{tab:imagenet-system}
\vspace{-1.5em}
\end{table}

\subsection{Results}
\label{subsec:imagenet-results}
\paragraph{ImageNet-1K.} Table \ref{tab:imagenet-system} (upper) shows the result comparison with two recent Transformer variants, DeiT \cite{Touvron2020} and Swin Transformers \cite{Liu2021swin}, as well as two ConvNets from architecture search - RegNets~\cite{Radosavovic2020designing}, EfficientNets~\cite{Tan2019efficientnet} and EfficientNetsV2~\cite{tan2021efficientnetv2}. \cnn{} competes favorably with two strong ConvNet baselines (RegNet~\cite{Radosavovic2020designing} and EfficientNet~\cite{Tan2019efficientnet}) in terms of the accuracy-computation trade-off, as well as the inference throughputs. \cnn{} also outperforms Swin Transformer of similar complexities \emph{across the board}, sometimes with a substantial margin (\eg 0.8\% for \cnn{}-T). Without specialized modules such as shifted windows or relative position bias, \cnn{s} also enjoy improved throughput compared to Swin Transformers.

A highlight from the results is \cnn{}-B at 384$^2$: it outperforms Swin-B by 0.6\% (85.1\% vs. 84.5\%), but with 12.5\% higher inference throughput (95.7 vs. 85.1 image/s). We note that the FLOPs/throughput advantage of \cnn{}-B over Swin-B becomes larger when the resolution increases from 224$^2$ to 384$^2$. Additionally, we observe an improved result of 85.5\% when further scaling to \cnn{}-L.

% \vspace{-5em}
\paragraph{ImageNet-22K.}
We present results with models fine-tuned from ImageNet-22K pre-training at Table \ref{tab:imagenet-system} (lower). 
These experiments are important since a widely held view is that vision Transformers have fewer inductive biases thus can perform better than ConvNets when pre-trained on a larger scale. 
 Our results demonstrate that properly designed ConvNets are \emph{not} inferior to vision Transformers when pre-trained with large dataset --- \cnn{s} still perform on par or better than similarly-sized Swin Transformers, with slightly higher throughput. Additionally, our \cnn{}-XL model achieves an accuracy of 87.8\% --- a decent improvement over \cnn{}-L at 384$^2$, demonstrating that \cnn{}s are scalable architectures. 
 
On ImageNet-1K, EfficientNetV2-L, a searched architecture equipped with advanced modules (such as Squeeze-and-Excitation~\cite{hu2018squeeze}) and progressive training procedure achieves top performance. However, with ImageNet-22K pre-training, ConvNeXt is able to outperform EfficientNetV2, further demonstrating the importance of large-scale training.

In Appendix~\ref{sec:robustness}, we discuss robustness and out-of-domain generalization results for \cnn{}.

\subsection{Isotropic \cnn{} \vs ViT}
\label{subsec:isotropic} 
In this ablation, we examine if our \cnn{} block design is generalizable to ViT-style~\cite{Dosovitskiy2021} isotropic architectures which have no downsampling layers and keep the same feature resolutions (\eg 14$\times$14) at all depths. We construct isotropic \cnn{}-S/B/L using the same feature dimensions as ViT-S/B/L (384/768/1024). Depths are set at 18/18/36 to match the number of parameters and FLOPs. The block structure remains the same (Fig.~\ref{fig:block}).
We use the supervised training results from DeiT \cite{Touvron2020} for ViT-S/B and MAE~\cite{he2021masked} for ViT-L, as they employ improved training procedures over the original ViTs \cite{Dosovitskiy2021}. \cnn{} models are trained with the same settings as before, but with longer warmup epochs. Results for ImageNet-1K at 224$^2$ resolution are in Table \ref{tab:non-hie}. We observe \cnn{} can perform generally on par with ViT, showing that our \cnn{} block design is competitive when used in non-hierarchical models. 

%##################################################################################################
\begin{table}[b]
\tablestyle{2pt}{1.0}
% \hspace{-1em}
\begin{tabular}{lcccccc}
 model & \#param. & FLOPs & \begin{tabular}[c]{@{}c@{}} throughput\\ (image / s) \end{tabular} & \begin{tabular}[c]{@{}c@{}} training \\ mem. (GB) \end{tabular} &  \begin{tabular}[c]{@{}c@{}}IN-1K \\ acc.\end{tabular} & \\
\shline
\vb ViT-S                         & 22M & 4.6G   &  978.5 & 4.9 & 79.8 \\
\gr
\cb \cnn{}-S (\textit{iso.})      & 22M & 4.3G   &  1038.7& 4.2 & 79.7 \\
\vb ViT-B                         & 87M & 17.6G  &  302.1 & 9.1 & 81.8 \\
\gr
\cb \cnn{}-B (\textit{iso.})      & 87M & 16.9G  &  320.1 & 7.7 & 82.0 \\
\vb ViT-L                         & 304M& 61.6G  &  93.1  & 22.5 & 82.6 \\
\gr
\cb \cnn{}-L (\textit{iso.})      & 306M& 59.7G  &  94.4  & 20.4 & 82.6 \\
\end{tabular}
\caption{\textbf{Comparing isotropic \cnn{} and ViT.} Training memory is measured on V100 GPUs with 32 per-GPU batch size.}
\label{tab:non-hie}
\end{table}
%##################################################################################################

\section{Empirical Evaluation on Downstream Tasks}
\paragraph{Object detection and segmentation on COCO.}

We fine-tune Mask R-CNN~\cite{He2017} and Cascade Mask R-CNN~\cite{Cai2018} on the COCO dataset with \cnn{} backbones. Following Swin Transformer~\cite{Liu2021swin}, we use multi-scale training, AdamW optimizer, and a 3$\times$ schedule. Further details and hyper-parameter settings can be found in Appendix~\ref{subsec:downstream-setting}.

Table~\ref{tab:coco} shows object detection and instance segmentation results comparing Swin Transformer, \cnn{}, and traditional ConvNet such as ResNeXt. Across different model complexities, \cnn{} achieves on-par or better performance than Swin Transformer. When scaled up to bigger models (\cnn{}-B/L/XL) pre-trained on ImageNet-22K, in many cases \emph{ConvNeXt is significantly better} (\eg +1.0 AP) than Swin Transformers in terms of box and mask AP.

\begin{table}
\tablestyle{6pt}{1.1}
\addtolength{\tabcolsep}{-4.5pt}
\vspace{2ex}
\scalebox{0.95}{
\begin{tabular}{@{}lcccccccc@{}}
backbone & FLOPs & FPS & $\text{AP}^{\text{box}}$ & $\text{AP}^{\text{box}}_{50}$ & $\text{AP}^{\text{box}}_{75}$ & $\text{AP}^{\text{mask}}$ & $\text{AP}^{\text{mask}}_{\text{50}}$ & $\text{AP}^{\text{mask}}_{75}$  \\
\shline
\multicolumn{9}{c}{\scriptsize{Mask-RCNN 3$\times$ schedule}} \\
\vb Swin-T      & 267G & 23.1     & 46.0 & 68.1 & 50.3 & 41.6 & 65.1 & 44.9 \\
\gr
\cb \cnn{}-T    & 262G & 25.6     & \textbf{46.2} & 67.9 & 50.8 & \textbf{41.7} & 65.0 & 44.9 \\
\hline
\multicolumn{9}{c}{\scriptsize{Cascade Mask-RCNN 3$\times$ schedule}} \\
\cb ResNet-50            & 739G & 16.2     & 46.3 & 64.3 & 50.5 & 40.1 & 61.7 & 43.4 \\
\cb X101-32              & 819G & 13.8  & 48.1 & 66.5 & 52.4 & 41.6 & 63.9 & 45.2 \\
\cb X101-64              & 972G & 12.6  & 48.3 & 66.4 & 52.3 & 41.7 & 64.0 & 45.1 \\
\Xhline{0.3\arrayrulewidth}
\vb Swin-T               & 745G & 12.2     & 50.4 & 69.2 & 54.7 & 43.7 & 66.6 & 47.3 \\
\gr
\cb \cnn{}-T             & 741G & 13.5     & \textbf{50.4} & 69.1 & 54.8 & \textbf{43.7} & 66.5 & 47.3 \\
\vb Swin-S               & 838G & 11.4     & 51.9	& 70.7	& 56.3	& 45.0	& 68.2	& 48.8 \\
\gr
\cb \cnn{}-S             & 827G & 12.0     & \textbf{51.9} & 70.8 & 56.5 & \textbf{45.0} & 68.4 & 49.1 \\
\vb Swin-B               & 982G & 10.7     & 51.9 & 70.5 & 56.4 & 45.0 & 68.1 & 48.9 \\
\gr
\cb \cnn{}-B             & 964G & 11.4     & \textbf{52.7} & 71.3 & 57.2 & \textbf{45.6} & 68.9 & 49.5 \\
\Xhline{0.3\arrayrulewidth}
\vb Swin-B$^\ddag$               & 982G & 10.7     &  53.0	& 71.8 & 57.5 & 45.8 & 69.4 & 49.7 \\
\gr
\cb \cnn{}-B$^\ddag$             & 964G & 11.5     & \textbf{54.0} & 73.1 & 58.8	& \textbf{46.9}	& 70.6 & 51.3 \\
\vb Swin-L$^\ddag$       & 1382G & 9.2  & 53.9 & 72.4 & 58.8 & 46.7 & 70.1 & 50.8 \\
\gr
\cb \cnn{}-L$^\ddag$     & 1354G & 10.0     & \textbf{54.8} & 73.8 & 59.8 & \textbf{47.6} & 71.3 & 51.7 \\
\gr
\cb \cnn{}-XL$^\ddag$     & 1898G & 8.6     & \textbf{55.2} & 74.2 & 59.9 & \textbf{47.7} & 71.6 & 52.2\\
\end{tabular}
} 
\caption[caption]{\textbf{COCO object detection and segmentation results} using Mask-RCNN and Cascade Mask-RCNN. $^\ddag$ indicates that the model is pre-trained on ImageNet-22K. ImageNet-1K pre-trained Swin results are from their Github repository~\cite{swindetcode}. AP numbers of the ResNet-50 and X101 models are from \cite{Liu2021swin}. We measure FPS on an A100 GPU. FLOPs are calculated with image size (1280, 800).
\label{tab:coco}
}
\end{table}

\paragraph{Semantic segmentation on ADE20K.}
We also evaluate \cnn{} backbones on the ADE20K semantic segmentation task with UperNet~\cite{Xiao2018}. All model variants are trained for 160K iterations with a batch size of 16. Other experimental settings follow~\cite{Bao2021} (see Appendix~\ref{subsec:downstream-setting} for more details). In Table \ref{tab:seg}, we report validation mIoU with multi-scale testing. \cnn{} models can achieve competitive performance across different model capacities, further validating the effectiveness of our architecture design. 

\begin{table}
    \centering
    \small
\addtolength{\tabcolsep}{-2.1pt}
\vspace{2ex}
\begin{tabular}{lcccc}
backbone & input crop. & mIoU & \#param. & FLOPs  \\
\Xhline{1.0pt}
\multicolumn{5}{c}{\scriptsize{ImageNet-1K pre-trained}} \\
\vb Swin-T & 512$^2$ & 45.8 & 60M & 945G  \\
\gr
\cb \cnn{}-T &  512$^2$ & \textbf{46.7} & 60M & 939G  \\
\vb Swin-S &  512$^2$ & 49.5 & 81M & 1038G  \\
\gr
\cb \cnn{}-S &  512$^2$ & \textbf{49.6} & 82M & 1027G  \\
\vb Swin-B &  512$^2$ & 49.7 & 121M & 1188G  \\
\gr
\cb \cnn{}-B &  512$^2$ & \textbf{49.9} & 122M & 1170G  \\
\hline
\multicolumn{4}{c}{\scriptsize{ImageNet-22K pre-trained}} \\
\vb Swin-B$^\ddag$ &  640$^2$ & 51.7 & 121M & 1841G  \\
\gr
\cb \cnn{}-B$^\ddag$ & 640$^2$ & \textbf{53.1} & 122M & 1828G  \\
\vb Swin-L$^\ddag$ & 640$^2$ & 53.5 & 234M & 2468G  \\
\gr
\cb \cnn{}-L$^\ddag$ & 640$^2$ & \textbf{53.7} & 235M & 2458G  \\
\gr
\cb \cnn{}-XL$^\ddag$ & 640$^2$ & \textbf{54.0} &  391M & 3335G  \\
\Xhline{1.0pt}
\end{tabular}
\vspace{1ex}
    \caption[caption]{\textbf{ADE20K validation results} using UperNet~\cite{Xiao2018}. $^\ddag$ indicates IN-22K  pre-training. Swins' results are from its GitHub repository~\cite{swincode}. Following Swin, we report mIoU results with multi-scale testing. FLOPs are based on input sizes of (2048, 512) and (2560, 640) for IN-1K and IN-22K pre-trained models, respectively.}
    \label{tab:seg}
    \normalsize
\end{table}

\paragraph{Remarks on model efficiency.} Under similar FLOPs, models with depthwise convolutions are known to be slower and consume more memory than ConvNets with only dense convolutions. It is natural to ask whether the design of \cnn{} will render it practically inefficient. As demonstrated throughout the paper, the inference throughputs of \cnn{s} are comparable to or exceed that of Swin Transformers. This is true for both classification and other tasks requiring higher-resolution inputs (see Table~\ref{tab:imagenet-system},\ref{tab:coco} for comparisons of throughput/FPS). Furthermore, we notice that training \cnn{s} requires less memory than training Swin Transformers. For example, training Cascade Mask-RCNN using \cnn{}-B backbone consumes 17.4GB of peak memory with a per-GPU batch size of 2, while the reference number for Swin-B is 18.5GB.
In comparison to vanilla ViT, both \cnn{} and Swin Transformer exhibit a more favorable accuracy-FLOPs trade-off due to the local computations. It is worth noting that this improved efficiency is a result of the \emph{ConvNet inductive bias}, and is not directly related to the self-attention mechanism in vision Transformers.

\section{Related Work}
\paragraph{Hybrid models.} 
In both the pre- and post-ViT eras, the hybrid model combining convolutions and self-attentions has been actively studied. 
Prior to ViT, the focus was on augmenting a ConvNet with self-attention/non-local modules~\cite{Wang2018,bello2019attention,srinivas2021bottleneck,ramachandran2019stand} to capture long-range dependencies.
The original ViT~\cite{Dosovitskiy2021} first studied a hybrid configuration, and a large body of follow-up works focused on reintroducing convolutional priors to ViT, either in an explicit~\cite{wu2021cvt, xu2021co, d2021convit, dai2021coatnet,Xiao2021, fan2021multiscale} or implicit~\cite{Liu2021swin} fashion.

\paragraph{Recent convolution-based approaches.} Han \emph{et al.}~\cite{han2021demystifying} show that local Transformer attention is equivalent to inhomogeneous dynamic depthwise conv. The MSA block in Swin is then replaced with a dynamic or regular depthwise convolution, achieving comparable performance to Swin. A concurrent work ConvMixer~\cite{convmixer} demonstrates that, in small-scale settings, depthwise convolution can be used as a promising mixing strategy. ConvMixer uses a smaller patch size to achieve the best results, making the throughput much lower than other baselines. GFNet~\cite{rao2021global} adopts Fast Fourier Transform (FFT) for token mixing. FFT is also a form of convolution, but with a global kernel size and circular padding. Unlike many recent Transformer or ConvNet designs, one primary goal of our study is to provide an in-depth look at the process of modernizing a standard ResNet and achieving state-of-the-art performance.

\section{Conclusions}
In the 2020s, vision Transformers, particularly hierarchical ones such as Swin Transformers, began to overtake ConvNets as the favored choice for generic vision backbones. The widely held belief is that vision Transformers are more accurate, efficient, and scalable than ConvNets. We propose \cnn{s}, a pure ConvNet model that can compete favorably with state-of-the-art hierarchical vision Transformers across multiple computer vision benchmarks, while retaining the simplicity and efficiency of standard ConvNets. In some ways, our observations are surprising while our \cnn{} model itself is not completely new --- many design choices have all been examined separately over the last decade, but not collectively. We hope that the new results reported in this study will challenge several widely held views and prompt people to rethink the importance of convolution in computer vision.

\paragraph{Acknowledgments.} We thank Kaiming He, Eric Mintun, Xingyi Zhou, Ross Girshick, and Yann LeCun for valuable discussions and feedback.

\appendix
\section*{\Large{Appendix}}
In this Appendix, we provide further experimental details (\S \ref{sec:setting}), robustness evaluation results (\S \ref{sec:robustness}), more modernization experiment results (\S\ref{sec:modernizing_result}), and a detailed network specification (\S \ref{sec:arch}). We further benchmark model throughput on A100 GPUs (\S \ref{sec:a100}). Finally, we discuss the limitations (\S \ref{sec:limit}) and societal impact (\S \ref{sec:impact}) of our work.

\section{Experimental Settings}
\label{sec:setting}
\subsection{ImageNet (Pre-)training} 
\label{subsec:setting}
We provide \cnn{s}' ImageNet-1K training and ImageNet-22K pre-training settings in Table~\ref{tab:train_detail}. The settings are used for our main results in Table~\ref{tab:imagenet-system} (Section~\ref{subsec:imagenet-results}).  All \cnn{} variants use the same setting, except the stochastic depth rate is customized for model variants. 
% For weight initialization, we use a truncated normal distribution with 0.2 standard deviation.

For experiments in ``modernizing a ConvNet'' (Section~\ref{sec:modernizing}), we also use Table~\ref{tab:train_detail}'s setting for ImageNet-1K, except EMA is disabled, as we find using EMA severely hurts models with BatchNorm layers.

For isotropic \cnn{}s (Section~\ref{subsec:isotropic}), the setting for ImageNet-1K in Table~\ref{sec:setting} is also adopted, but warmup is extended to 50 epochs, and layer scale is disabled for isotropic \cnn{}-S/B. The stochastic depth rates are 0.1/0.2/0.5 for isotropic \cnn{}-S/B/L.

\begin{table}[h]
\tablestyle{5.0pt}{1.02}
\footnotesize
\begin{tabular}{@{\hskip -0.05ex}l|c@{\hskip 1ex}c}
& \cnn{}-T/S/B/L & \cnn{}-T/S/B/L/XL \\
\multirow{2}{*}{(pre-)training config} & ImageNet-1K & ImageNet-22K \\
& 224$^2$ & 224$^2$ \\
\shline
weight init & trunc. normal (0.2) & trunc. normal (0.2) \\
optimizer & AdamW & AdamW\\
base learning rate & 4e-3 & 4e-3 \\
weight decay & 0.05 & 0.05 \\
optimizer momentum & $\beta_1, \beta_2{=}0.9, 0.999$ & $\beta_1, \beta_2{=}0.9, 0.999$ \\
batch size & 4096 & 4096 \\
training epochs & 300 & 90 \\
learning rate schedule & cosine decay & cosine decay \\
warmup epochs & 20 & 5 \\
warmup schedule & linear & linear \\
layer-wise lr decay \cite{Clark2020,Bao2021} & None & None \\
randaugment \cite{Cubuk2020} & (9, 0.5) & (9, 0.5) \\
mixup \cite{Zhang2018a} & 0.8 & 0.8 \\
cutmix \cite{Yun2019} & 1.0 & 1.0 \\
random erasing \cite{Zhong2020} & 0.25 & 0.25 \\
label smoothing \cite{Szegedy2016a} & 0.1 & 0.1 \\
stochastic depth \cite{Huang2016deep} & 0.1/0.4/0.5/0.5 & 0.0/0.0/0.1/0.1/0.2 \\
layer scale \cite{Touvron2021GoingDW} & 1e-6 & 1e-6 \\
head init scale \cite{Touvron2021GoingDW} & None & None \\
gradient clip & None & None \\
exp. mov. avg. (EMA) \cite{Polyak1992} & 0.9999 & None\\

\end{tabular}
\caption{\textbf{ImageNet-1K/22K (pre-)training settings}. Multiple stochastic depth rates (e.g., 0.1/0.4/0.5/0.5) are for each model (e.g., \cnn{}-T/S/B/L) respectively.}
\label{tab:train_detail}
\end{table}

\begin{table}[h]
\tablestyle{5.0pt}{1.02}
\footnotesize
\begin{tabular}{@{\hskip -0.05ex}l@{\hskip 2.6ex}|cc}
 &  \cnn{-B/L}  & \cnn{-T/S/B/L/XL}  \\
\multirow{2}{*}{pre-training config} & ImageNet-1K   & ImageNet-22K  \\
 &  224$^2$  & 224$^2$ \\
\hline
\multirow{2}{*}{fine-tuning config} & ImageNet-1K   & ImageNet-1K  \\
 & 384$^2$ & 224$^2$ and 384$^2$ \\
\shline
optimizer & AdamW & AdamW\\
base learning rate & 5e-5 & 5e-5 \\
weight decay & 1e-8 & 1e-8 \\
optimizer momentum & $\beta_1, \beta_2{=}0.9, 0.999$ & $\beta_1, \beta_2{=}0.9, 0.999$ \\
batch size & 512 & 512 \\
training epochs & 30 & 30 \\
learning rate schedule & cosine decay & cosine decay \\
layer-wise lr decay & 0.7 & 0.8  \\
warmup epochs & None & None \\
warmup schedule & N/A & N/A \\
randaugment & (9, 0.5) & (9, 0.5) \\
mixup & None & None \\
cutmix  & None & None \\
random erasing & 0.25 & 0.25 \\
label smoothing  & 0.1 & 0.1 \\
stochastic depth  & 0.8/0.95 & 0.0/0.1/0.2/0.3/0.4 \\
layer scale & pre-trained & pre-trained \\
head init scale & 0.001 & 0.001 \\
gradient clip & None & None \\
exp. mov. avg. (EMA) & None & None(T-L)/0.9999(XL) \\
\end{tabular}
\caption{\textbf{ImageNet-1K fine-tuning settings}. Multiple values (e.g., 0.8/0.95) are for each model (e.g., \cnn{}-B/L) respectively. }
\label{tab:ft_detail}
\end{table}

\subsection{ImageNet Fine-tuning} 
\label{subsec:ft-setting}
We list the settings for fine-tuning on ImageNet-1K in Table~\ref{tab:ft_detail}. The fine-tuning starts from the final model weights obtained in pre-training, without using the EMA weights, even if in pre-training EMA is used and EMA accuracy is reported. This is because we do not observe improvement if we fine-tune with the EMA weights (consistent with observations in \cite{Touvron2020}). The only exception is \cnn{}-L pre-trained on ImageNet-1K, where the model accuracy is significantly lower than the EMA accuracy due to overfitting, and we select its best EMA model during pre-training as the starting point for fine-tuning.

In fine-tuning, we use layer-wise learning rate decay~\cite{Clark2020,Bao2021} with every 3 consecutive blocks forming a group. When the model is fine-tuned at 384$^2$ resolution, we use a crop ratio of 1.0 (i.e., no cropping) during testing following \cite{rw2019timm,Touvron2021GoingDW,swincode}, instead of 0.875 at 224$^2$.

\subsection{Downstream Tasks}
\label{subsec:downstream-setting}

For ADE20K and COCO experiments, we follow the training settings used in BEiT~\cite{Bao2021} and Swin~\cite{Liu2021swin}. We also use MMDetection~\cite{mmdetection} and MMSegmentation~\cite{mmseg2020} toolboxes. We use the final model weights (instead of EMA weights) from ImageNet pre-training as network initializations. 

We conduct a lightweight sweep for COCO experiments including learning rate \{1e-4, 2e-4\}, layer-wise learning rate decay~\cite{Bao2021} \{0.7, 0.8, 0.9, 0.95\}, and stochastic depth rate \{0.3, 0.4, 0.5, 0.6, 0.7, 0.8\}. We fine-tune the ImageNet-22K pre-trained Swin-B/L on COCO using the same sweep. We use the official code and pre-trained model weights~\cite{swindetcode}.

The hyperparameters we sweep for ADE20K experiments include learning rate \{8e-5, 1e-4\}, layer-wise learning rate decay \{0.8, 0.9\}, and stochastic depth rate \{0.3, 0.4, 0.5\}. We report validation mIoU results using multi-scale testing. Additional single-scale testing results are in Table~\ref{tab:seg-ss}.

\begin{table}[h]
    \centering
    \small
\addtolength{\tabcolsep}{-2.1pt}
\begin{tabular}{lcc}
backbone & input crop. & mIoU \\
\Xhline{1.0pt}
\multicolumn{3}{c}{\scriptsize{ImageNet-1K pre-trained}} \\
\cb \cnn{}-T &  512$^2$ & {46.0} \\
\cb \cnn{}-S &  512$^2$ & {48.7}  \\
\cb \cnn{}-B &  512$^2$ & {49.1}  \\
\hline
\multicolumn{3}{c}{\scriptsize{ImageNet-22K pre-trained}} \\
\cb \cnn{}-B$^\ddag$ & 640$^2$ & {52.6}  \\
\cb \cnn{}-L$^\ddag$ & 640$^2$ & {53.2}  \\
\cb \cnn{}-XL$^\ddag$ & 640$^2$ & {53.6}  \\
\Xhline{1.0pt}
\end{tabular}
    \caption[caption]{\textbf{ADE20K validation results} with single-scale testing.}
    \label{tab:seg-ss}
    \normalsize
\end{table}

\section{Robustness Evaluation}
\label{sec:robustness}
Additional robustness evaluation results for ConvNeXt models are presented in Table~\ref{tab:robustness}. We directly test our ImageNet-1K trained/fine-tuned classification models on several robustness benchmark datasets such as ImageNet-A~\cite{hendrycks2021natural}, ImageNet-R~\cite{hendrycks2021many}, ImageNet-Sketch~\cite{wang2019learning} and ImageNet-C/$\bar{\text{C}}$~\cite{hendrycks2018benchmarking, mintun2021interaction} datasets. We report mean corruption error (mCE) for ImageNet-C, corruption error for ImageNet-$\bar{\text{C}}$, and top-1 Accuracy for all other datasets. 

ConvNeXt (in particular the large-scale model variants) exhibits promising robustness behaviors, outperforming state-of-the-art robust transformer models~\cite{mao2021towards} on several benchmarks. With extra ImageNet-22K data, ConvNeXt-XL demonstrates strong domain generalization capabilities (\eg achieving 69.3\%/68.2\%/55.0\% accuracy on ImageNet-A/R/Sketch benchmarks, respectively). We note that these robustness evaluation results were acquired without using any specialized modules or additional fine-tuning procedures.
 
\begin{table}[h]
\tablestyle{7.3pt}{1.1}
\addtolength{\tabcolsep}{-4.5pt}
\scalebox{0.86}{
\begin{tabular}{llcccccccc}
        Model & Data/Size & FLOPs / Params & Clean & C ($\downarrow$) & $\bar{\text{C}}$ ($\downarrow$) & A & R & SK\\
        \midrule
        ResNet-50 & 1K/224$^2$ & 4.1 / 25.6 & 76.1 & 76.7 & 57.7 & 0.0 & 36.1 & 24.1 \\
        \midrule
        Swin-T~\cite{Liu2021swin} & 1K/224$^2$ &  4.5 / 28.3 & 81.2 & 62.0 & - & 21.6 & 41.3 & 29.1 \\
        RVT-S*~\cite{mao2021towards} & 1K/224$^2$ &  4.7 / 23.3 & 81.9 & 49.4 & 37.5 & 25.7 & 47.7 & 34.7  \\
        \gr
        {ConvNeXt-T} & 1K/224$^2$ &  4.5 / 28.6 & 82.1 & 53.2 & 40.0 & 24.2 & 47.2 & 33.8  \\
        Swin-B~\cite{Liu2021swin}  & 1K/224$^2$ &  15.4 / 87.8 & 83.4 & 54.4 & - & 35.8 & 46.6 & 32.4 \\
        RVT-B*~\cite{mao2021towards} & 1K/224$^2$ & 17.7 / 91.8 & 82.6 & 46.8 & \textbf{30.8} & 28.5 & 48.7 & 36.0    \\
        \gr
        {ConvNeXt-B} & 1K/224$^2$ & 15.4 / 88.6 & \textbf{83.8} & \textbf{46.8} & 34.4 & \textbf{36.7} & \textbf{51.3} & \textbf{38.2}  \\
        \midrule
        \gr
        {ConvNeXt-B} & 22K/384$^2$ & 45.1 / 88.6 & 86.8 & 43.1 & 30.7 & 62.3 & 64.9 & 51.6 \\
        \gr
        {ConvNeXt-L} & 22K/384$^2$ & 101.0 / 197.8 & 87.5 & 40.2 & 29.9 & 65.5 & 66.7 & 52.8 \\
        \gr
        {ConvNeXt-XL} & 22K/384$^2$ &  179.0 / 350.2 & \textbf{87.8} & \textbf{38.8} & \textbf{27.1} & \textbf{69.3} & \textbf{68.2} & \textbf{55.0}  \\
    \end{tabular}
} 
\caption[caption]{\textbf{Robustness evaluation of ConvNeXt}. We do not make use of any specialized modules or additional fine-tuning procedures. \label{tab:robustness}}
\vspace{-3ex}
\end{table}
%##################################################################################################

\begin{table*}[t]
\centering
\addtolength{\tabcolsep}{-2pt}
\vspace{4ex}
\scalebox{0.85}{
\begin{tabular}{c|c|c|c|c}
& \begin{tabular}[c]{@{}c@{}} output size\end{tabular} & \cb ResNet-50 & \cb \cnn{}-T  & \vb Swin-T \\
\hline
\multirow{2}{*}{stem} & \multirow{2}{*}{\begin{tabular}[c]{@{}c@{}} 56$\times$56 \end{tabular}} & 
7$\times$7, 64, stride 2  
& \multirow{2}{*}{4$\times$4, 96, stride 4} & \multirow{2}{*}{4$\times$4, 96, stride 4} \\ 

& & 3$\times$3 max pool, stride 2 & & \\

\hline
\multirow{5}{*}{res2} & 
\multirow{5}{*}{\begin{tabular}[c]{@{}c@{}} 56$\times$56 \end{tabular}} & 
\multirow{5}{*}{$\begin{bmatrix}\text{1$\times$1, 64}\\\text{3$\times$3, 64}\\\text{1$\times$1, 256}\end{bmatrix}$ $\times$ 3}  & 
\multirow{5}{*}{$\begin{bmatrix}\text{d7$\times$7, 96}\\\text{1$\times$1, 384}\\\text{1$\times$1, 96}\end{bmatrix}$ $\times$ 3} &
\multirow{5}{*}{$\begin{matrix}\begin{bmatrix}\text{1$\times$1, 96$\times$3}\\\text{MSA, w7$\times$7, H=3, rel. pos.}\\\text{1$\times$1, 96}\end{bmatrix}\\\begin{bmatrix}\text{\quad1$\times$1, 384\quad}\\\text{\quad1$\times$1, 96\quad}\end{bmatrix}\end{matrix}$  $\times$ 2} \\
& & & & \\
& & & & \\
& & & & \\
& & & & \\
\hline
\multirow{5}{*}{res3} & 
\multirow{5}{*}{\begin{tabular}[c]{@{}c@{}} 28$\times$28 \end{tabular}} & 
\multirow{5}{*}{$\begin{bmatrix}\text{1$\times$1, 128}\\\text{3$\times$3, 128}\\\text{1$\times$1, 512}\end{bmatrix}$ $\times$ 4}  & 
\multirow{5}{*}{$\begin{bmatrix}\text{d7$\times$7, 192}\\\text{1$\times$1, 768}\\\text{1$\times$1, 192}\end{bmatrix}$ $\times$ 3} &
\multirow{5}{*}{$\begin{matrix}\begin{bmatrix}\text{1$\times$1, 192$\times$3}\\\text{MSA, w7$\times$7, H=6, rel. pos.}\\\text{1$\times$1, 192}\end{bmatrix}\\\begin{bmatrix}\text{\quad1$\times$1, 768\quad}\\\text{\quad1$\times$1, 192\quad}\end{bmatrix}\end{matrix}$  $\times$ 2} \\
& & & & \\
& & & & \\
& & & & \\
& & & & \\
\hline
\multirow{5}{*}{res4} & 
\multirow{5}{*}{\begin{tabular}[c]{@{}c@{}} 14$\times$14 \end{tabular}} & 
\multirow{5}{*}{$\begin{bmatrix}\text{1$\times$1, 256}\\\text{3$\times$3, 256}\\\text{1$\times$1, 1024}\end{bmatrix}$ $\times$ 6}  & 
\multirow{5}{*}{$\begin{bmatrix}\text{d7$\times$7, 384}\\\text{1$\times$1, 1536}\\\text{1$\times$1, 384}\end{bmatrix}$ $\times$ 9} &
\multirow{5}{*}{$\begin{matrix}\begin{bmatrix}\text{1$\times$1, 384$\times$3}\\\text{MSA, w7$\times$7, H=12, rel. pos.}\\\text{1$\times$1, 384}\end{bmatrix}\\\begin{bmatrix}\text{\quad1$\times$1, 1536\quad}\\\text{\quad1$\times$1, 384\ \quad}\end{bmatrix}\end{matrix}$  $\times$ 6} \\
& & & & \\
& & & & \\
& & & & \\
& & & & \\
\hline
\multirow{5}{*}{res5} & 
\multirow{5}{*}{\begin{tabular}[c]{@{}c@{}} 7$\times$7 \end{tabular}} & 
\multirow{5}{*}{$\begin{bmatrix}\text{1$\times$1, 512}\\\text{3$\times$3, 512}\\\text{1$\times$1, 2048}\end{bmatrix}$ $\times$ 3}  & 
\multirow{5}{*}{$\begin{bmatrix}\text{d7$\times$7, 768}\\\text{1$\times$1, 3072}\\\text{1$\times$1, 768}\end{bmatrix}$ $\times$ 3} &
\multirow{5}{*}{$\begin{matrix}\begin{bmatrix}\text{1$\times$1, 768$\times$3}\\\text{MSA, w7$\times$7, H=24, rel. pos.}\\\text{1$\times$1, 768}\end{bmatrix}\\\begin{bmatrix}\text{\quad1$\times$1, 3072\quad}\\\text{\quad1$\times$1, 768\ \quad}\end{bmatrix}\end{matrix}$  $\times$ 2} \\
& & & & \\
& & & & \\
& & & & \\
& & & & \\
\hline
\multicolumn{2}{c|}{FLOPs}
&
$4.1 \times 10^9$
&
$4.5 \times 10^9$
&
$4.5 \times 10^9$
\\
\hline
\multicolumn{2}{c|}{\# params.}
&
$25.6 \times 10^6$
&
$28.6 \times 10^6$
&
$28.3 \times 10^6$ \\
\hline

\end{tabular}
}
\normalsize
\caption{\textbf{Detailed architecture specifications} for ResNet-50, \cnn{}-T and Swin-T.}
\label{table:arch-spec}
\end{table*}

\section{Modernizing ResNets: detailed results}
\label{sec:modernizing_result}
Here we provide detailed tabulated results for the \emph{modernization} experiments, at both ResNet-50 / Swin-T and ResNet-200 / Swin-B regimes. The ImageNet-1K top-1 accuracies and FLOPs for each step are shown in Table~\ref{tab:modernizing-t} and~\ref{tab:modernizing-b}. ResNet-50 regime experiments are run with 3 random seeds.

For ResNet-200, the initial number of blocks at each stage is (3, 24, 36, 3). We change it to Swin-B's (3, 3, 27, 3) at the step of changing stage ratio. This drastically reduces the FLOPs, so at the same time, we also increase the width from 64 to 84 to keep the FLOPs at a similar level. After the step of adopting depthwise convolutions, we further increase the width to 128 (same as Swin-B's) as a separate step. 

The observations on the ResNet-200 regime are mostly consistent with those on ResNet-50 as described in the main paper. One interesting difference is that inverting dimensions brings a larger improvement at ResNet-200 regime than at ResNet-50 regime (+0.79\% \vs +0.14\%). The performance gained by increasing kernel size also seems to saturate at kernel size 5 instead of 7. Using fewer normalization layers also has a bigger gain compared with the ResNet-50 regime (+0.46\% \vs +0.14\%).

\begin{table}[tp]
\centering
\small
\begin{tabular}{lcc}
model & IN-1K acc. & GFLOPs \\
\shline
\gr
ResNet-50 (PyTorch\cite{torchvision}) & 76.13 &4.09 \\
\gr
ResNet-50 (enhanced recipe)     & 78.82 $\pm$ 0.07 &4.09 \\
stage ratio                     & 79.36 $\pm$ 0.07 &4.53 \\
``patchify'' stem               & 79.51 $\pm$ 0.18 &4.42 \\
depthwise conv                  & 78.28 $\pm$ 0.08 &2.35 \\
increase width                  & 80.50 $\pm$ 0.02 &5.27 \\
inverting dimensions            & 80.64 $\pm$ 0.03 &4.64 \\
move up depthwise conv          & 79.92 $\pm$ 0.08 &4.07 \\
kernel size $\rightarrow$ 5     & 80.35 $\pm$ 0.08 &4.10 \\
kernel size $\rightarrow$ 7     & 80.57 $\pm$ 0.14 &4.15 \\
kernel size $\rightarrow$ 9     & 80.57 $\pm$ 0.06 &4.21 \\
kernel size $\rightarrow$ 11    & 80.47 $\pm$ 0.11 &4.29 \\
ReLU $\rightarrow$ GELU         & 80.62 $\pm$ 0.14 &4.15 \\
fewer activations               & 81.27 $\pm$ 0.06 &4.15 \\
fewer norms                     & 81.41 $\pm$ 0.09 &4.15 \\
BN $\rightarrow$ LN             & 81.47 $\pm$ 0.09 &4.46 \\
\gr
separate d.s. conv (\cnn{}-T)     & 81.97 $\pm$ 0.06 &4.49 \\
\gr
Swin-T \cite{Liu2021swin}       &81.30 &4.50 \\
\shline
\end{tabular}
\caption{\textbf{Detailed results for modernizing a ResNet-50.} Mean and standard deviation are obtained by training the network with three different random seeds.}
\label{tab:modernizing-t}
\end{table}

\begin{table}[htp]
\centering
\small
\addtolength{\tabcolsep}{1.5pt}
\begin{tabular}{lcc}
model & IN-1K acc. & GFLOPs \\
\shline
\gr
ResNet-200 \cite{He2016a} & 78.20 & 15.01 \\
\gr
ResNet-200 (enhanced recipe) &81.14 &15.01 \\
stage ratio and increase width  &81.33 &14.52 \\
``patchify'' stem &81.59 &14.38 \\
depthwise conv &80.54 &7.23 \\
increase width  &81.85 &16.76 \\
inverting dimensions &82.64 &15.68 \\
move up depthwise conv &82.04 &14.63 \\
kernel size $\rightarrow$ 5 &82.32 &14.70 \\
kernel size $\rightarrow$ 7 &82.30 &14.81 \\
kernel size $\rightarrow$ 9 &82.27 &14.95 \\
kernel size $\rightarrow$ 11 &82.18 &15.13 \\
ReLU $\rightarrow$ GELU &82.19 &14.81 \\
fewer activations &82.71 &14.81 \\
fewer norms &83.17 &14.81 \\
BN $\rightarrow$ LN &83.35 &14.81 \\
\gr
separate d.s. conv (\cnn{}-B) &83.60 &15.35 \\
\gr
Swin-B\cite{Liu2021swin}  &83.50 &15.43 \\
\shline
\end{tabular}
\caption{\textbf{Detailed results for modernizing a ResNet-200.}}
\label{tab:modernizing-b}
\end{table}

%##################################################################################################

\section{Detailed Architectures}
\label{sec:arch}
We present a detailed architecture comparison between ResNet-50, \cnn{}-T and Swin-T in Table~\ref{table:arch-spec}. For differently sized \cnn{}s, only the number of blocks and the number of channels at each stage differ from \cnn{}-T (see Section~\ref{sec:convnext_config} for details). \cnn{}s enjoy the simplicity of standard ConvNets, but compete favorably with Swin Transformers in visual recognition.

%%##################################################################################################

\section{Benchmarking on A100 GPUs}
\label{sec:a100}

Following Swin Transformer~\cite{Liu2021swin}, the ImageNet models' inference throughputs in Table~\ref{tab:imagenet-system} are benchmarked using a V100 GPU, where \cnn{} is slightly faster in inference than Swin Transformer with a similar number of parameters. We now benchmark them on the more advanced A100 GPUs, which support the TensorFloat32 (TF32) tensor cores. We employ PyTorch~\cite{Paszke2019} version 1.10 to use the latest ``Channel Last'' memory layout~\cite{clpytorch} for further speedup.

We present the results in Table~\ref{tab:a100}. Swin Transformers and \cnn{}s both achieve faster inference throughput than V100 GPUs, but \cnn{}s' advantage is now significantly greater, sometimes \emph{up to 49\% faster}. This preliminary study shows promising signals that ConvNeXt, employed with standard ConvNet modules and simple in design, could be practically more efficient models on modern hardwares.

\newcommand{\spaceddash}{\ \ \ --  \ \ }
\begin{table}[ht]
\centering
\small
\addtolength{\tabcolsep}{-3.pt}
\vspace{1ex}
\begin{tabular}{lcccc}
   model & \begin{tabular}[c]{@{}c@{}}image \\ size\end{tabular} & FLOPs & \begin{tabular}[c]{@{}c@{}}throughput\\ (image / s)\end{tabular} & \begin{tabular}[c]{@{}c@{}} IN-1K / 22K \\ trained, 1K acc.\end{tabular} \\
\shline
\vb Swin-T   & 224$^2$ & 4.5G & 1325.6 & 81.3 /\spaceddash{} \\
\gr
\cb \cnn{}-T & 224$^2$ & 4.5G & \textbf{1943.5} \scriptsize{(+47\%)} & \textbf{82.1} /\spaceddash{} \\ % dp 0.1

\vb Swin-S   & 224$^2$ & 8.7G & 857.3 & 83.0 /\spaceddash{} \\
\gr
\cb \cnn{}-S & 224$^2$ & 8.7G & \textbf{1275.3} \scriptsize{(+49\%)} & \textbf{83.1} /\spaceddash{}  \\ % dp 0.4

\vb Swin-B   & 224$^2$ & 15.4G& 662.8 & 83.5 / 85.2\\
\gr
\cb \cnn{}-B & 224$^2$ & 15.4G& \textbf{969.0} \scriptsize{(+46\%)} & \textbf{83.8} / \textbf{85.8} \\ % dp 0.5

\vb Swin-B   & 384$^2$ & 47.1G& 242.5 & 84.5 / 86.4 \\
\gr
\cb \cnn{}-B & 384$^2$ & 45.0G& \textbf{336.6} \scriptsize{(+39\%)} & \textbf{85.1} / \textbf{86.8} \\

\vb Swin-L   & 224$^2$ & 34.5G & 435.9 &  \spaceddash{}/ 86.3         \\
\gr
\cb \cnn{}-L & 224$^2$ & 34.4G & \textbf{611.5} \scriptsize{(+40\%)} & \textbf{84.3} / \textbf{86.6} \\

\vb Swin-L   & 384$^2$ & 103.9G& 157.9 & \spaceddash{}/ 87.3          \\
\gr
\cb \cnn{}-L & 384$^2$ & 101.0G& \textbf{211.4} \scriptsize{(+34\%)} &  85.5 / \textbf{87.5}  \\
\gr

\gr
\cb \cnn{}-XL & 224$^2$ & 60.9G & \textbf{424.4}  &  \spaceddash{}/ \textbf{87.0}  \\
\gr
\cb \cnn{}-XL & 384$^2$ &179.0G & \textbf{147.4}  &  \spaceddash{}/ \textbf{87.8}  \\

\hline

\end{tabular}
\normalsize
\caption{\textbf{Inference throughput comparisons on an A100 GPU.} Using TF32 data format and ``channel last'' memory layout, \cnn{} enjoys up to $\sim$49\% higher throughput compared with a Swin Transformer with similar FLOPs.}
\label{tab:a100}
\end{table}

\section{Limitations}
\label{sec:limit}
We demonstrate ConvNeXt, a pure ConvNet model, can perform as good as a hierarchical vision Transformer on image classification, object detection, instance and semantic segmentation tasks. While our goal is to offer a broad range of evaluation tasks, we recognize computer vision applications are even more diverse. ConvNeXt may be more suited for certain tasks, while Transformers may be more flexible for others. A case in point is multi-modal learning, in which a cross-attention module may be preferable for modeling feature interactions across many modalities. Additionally, Transformers may be more flexible when used for tasks requiring discretized, sparse, or structured outputs. We believe the architecture choice should meet the needs of the task at hand while striving for simplicity.

\section{Societal Impact}
\label{sec:impact}
In the 2020s, research on visual representation learning began to place enormous demands on computing resources. While larger models and datasets improve performance across the board, they also introduce a slew of challenges. ViT, Swin, and ConvNeXt all perform best with their huge model variants. Investigating those model designs inevitably results in an increase in carbon emissions. One important direction, and a motivation for our paper, is to strive for simplicity --- with more sophisticated modules, the network's design space expands enormously, obscuring critical components that contribute to the performance difference. Additionally, large models and datasets present issues in terms of model robustness and fairness. 
Further investigation on the robustness behavior of ConvNeXt vs. Transformer will be an interesting research direction. In terms of data, our findings indicate that ConvNeXt models benefit from pre-training on large-scale datasets. While our method makes use of the publicly available ImageNet-22K dataset, individuals may wish to acquire their own data for pre-training. A more circumspect and responsible approach to data selection is required to avoid potential concerns with data biases.

{\small\bibliographystyle{ieee_fullname}\bibliography{main}}

\end{document}